\documentclass[letterpaper, 10 pt, conference]{ieeeconf} % Comment this line out if you need a4paper

\usepackage{mathtools}
\usepackage[listings,skins]{tcolorbox}
\usepackage{booktabs}
\usepackage{amssymb}

\usepackage{float}
\usepackage[inkscapearea=page]{svg}
\usepackage{calc}
\usepackage{xcolor}
\definecolor{hivePurple}{RGB}{102, 51, 153}
\usepackage{caption}
\usepackage{subcaption}
\usepackage{hyperref}
\hypersetup{
    colorlinks=true,
    linkcolor=blue,
    urlcolor=blue,
    citecolor=blue
}

 \usepackage[style=ieee, backend=bibtex, defernumbers=true]{biblatex}
\usepackage{multirow}
\usepackage{multicol}
\usepackage{amsthm}

\usepackage{colortbl}

\definecolor{gr}{rgb}{0.921, 0.972, 0.905}
\definecolor{pink}{rgb}{0.972, 0.905, 0.917}
\definecolor{redishh}{rgb}{0.9, 0.17, 0.31}
\definecolor{redish}{rgb}{1.0, 0.01, 0.24}
\definecolor{antique}{rgb}{0.57, 0.36, 0.51}
\definecolor{darkcandy}{rgb}{0.64, 0.0, 0.0}
\definecolor{pastel}{rgb}{0.09, 0.45, 0.27}

\IEEEoverridecommandlockouts        % This command is only needed if 
\title{
Scaling Public Health Text Annotation: Zero-Shot Learning vs. Crowdsourcing for Improved Efficiency and Labeling Accuracy
}

\author{Kamyar Kazari$^{1}$, Yong Chen$^{1}$, and Zahra Shakeri$^{1,2,3}$
    \thanks{$^{1}$ Kamyar Kazari is with the Institute of Health Policy, Management, and Evaluation (IHPME), Dalla Lana School of Public Health, University of Toronto, Canada. 
      {\tt\small k.kazari@mail.utoronto.ca}}%
    \thanks{$^{1}$ Yong Chen is with IHPME, Dalla Lana School of Public Health, University of Toronto, Canada.} %
\thanks{$^{1,2,3}$Zahra Shakeri is with IHPME; the Dalla Lana School of Public Health; the Faculty of Information; and the Schwartz Reisman Institute at the University of Toronto, Canada. 
    {\tt\small}}%
}

\begin{document}
\maketitle
\thispagestyle{empty}
\pagestyle{empty}

%%%%%%%%%%%%%%%%%%%%%%%%%%%%%%%%%%%%%%%%%%%%%%%%%%%%%%%%%%%%%%%%%%%%
\begin{abstract}

Public health researchers are increasingly interested in using social media data to study health-related behaviors, but manually labeling this data can be labor-intensive and costly. This study explores whether zero-shot labeling using large language models (LLMs) can match or surpass conventional crowd-sourced annotation for Twitter posts related to sleep disorders, physical activity, and sedentary behavior. Multiple annotation pipelines were designed to compare labels produced by domain experts, crowd workers, and LLM-driven approaches under varied prompt-engineering strategies. Our findings indicate that LLMs can rival human performance in straightforward classification tasks and significantly reduce labeling time, yet their accuracy diminishes for tasks requiring more nuanced domain knowledge. These results clarify the trade-offs between automated scalability and human expertise, demonstrating conditions under which LLM-based labeling can be efficiently integrated into public health research without undermining label quality.

\end{abstract}

%----------------------------------
\definecolor{amaranth}{rgb}{0.9, 0.17, 0.31}
\definecolor{gr}{rgb}{0.55, 0.71, 0.0}
\definecolor{ashgrey}{rgb}{0.7, 0.75, 0.71}
%----------------------------------

%%%%%%%%%%%%%%%%%%%%%%%%%%%%%%%%%%%%%%%%%%%%%%%%%%%%%%%%%%%%%%%%%%%%%%%%%%%%%%%%

\vspace{-1mm}
\section{INTRODUCTION}

Public health research often relies on large collections of text to study health-related behaviors and patterns. These texts can originate from diverse sources, including social media, where users share personal activities, symptoms, and opinions in real time. Efficiently labeling such unstructured data remains a major challenge, because the volume and variability of posts can exceed the capacity of traditional annotation workflows \cite{DAR2024112526}. Crowdsourcing platforms like Amazon Mechanical Turk (AMT) help distribute this workload among many workers \cite{shakeri2022crowdsourcing}, but high costs, oversight requirements, and prolonged turnaround times remain problematic, especially when tasks demand domain-specific expertise \cite{marshall2023broke,karpinska2021perils}.

Recent advances in large language models (LLMs) offer a compelling alternative to conventional text annotation. LLMs have shown strong performance in zero-shot and few-shot classification tasks, even on challenging datasets \cite{brown2020language,radford2019language,qin2023chatgptgeneralpurposenaturallanguage}. Some studies have used ChatGPT to generate labels with justifications \cite{törnberg2023chatgpt4outperformsexpertscrowd,zhu2023chatgptreproducehumangeneratedlabels}, though manual interactions limit scalability. Direct use of the OpenAI application programming interface (API) bypasses this restriction and enables fast processing of large datasets \cite{Gilardi_2023}. This paper proposes a framework for comparing GPT-4~Turbo, in a zero-shot setting, to AMT annotators on 12,000 posts from X (formerly Twitter). Each post concerns one of three health-related topics: physical activity, sedentary behavior, or sleep problems, labeled as \textquotesingle{}Yes\textquotesingle{}, \textquotesingle{}No\textquotesingle{}, or \textquotesingle{}Unclear\textquotesingle{}. Analysis centers on accuracy, cost, and time, all of which influence decisions in public health interventions.

One key observation is that language in domains like physical activity and sedentary behavior is more direct, allowing LLMs to match or surpass crowd workers. In contrast, tasks involving subtle or context-dependent information, such as sleep medications or cryptic references to nighttime routines, pose challenges that may be addressed only with domain-specific knowledge. Handling such tweets requires awareness of specialized terminology or cultural references often unfamiliar to a generic LLM. Our approach explores strategies to mitigate these hurdles, including refined prompts and batch labeling via the GPT-4~Turbo API, while also examining cost-saving methods for large-scale annotation projects.

Our findings suggest that a hybrid workflow, where automated labeling addresses clear-cut cases and human expertise oversees ambiguous or domain-specific tasks, can cut labeling budgets and accelerate project timelines. Public health projects that require rapid analysis of emerging topics, such as early detection of infectious disease outbreaks and environmental hazards surveillance, stand to benefit. This work offers a structured comparison between zero-shot LLM labeling and crowdsourced annotation, outlining scenarios where each approach excels, and contributes practical insights into integrating large-scale automation within health research pipelines. 

The rest of the paper is organized as follows: Section~II defines the problem and details the methodology. Section~III reports key results on accuracy, time, and cost. Section~V discusses challenges, potential solutions, and future directions, concluding with a summary of key findings.

\vspace{-1.5mm}
%%%%%%%%%%%%%%%%%%%%%%%%%%%%%%%%%%%%%%%%%%%%%%%%%%%%%%%%%%%%%%%%%%%%%%%%%%%%%%%%
\section{Methods}

We formalize our study using a mathematical framework to define classification tasks, data collection, annotation, and LLM-based labeling. We then detail data curation and preprocessing before outlining the model deployment pipeline, including LLM integration and performance evaluation.
\vspace{-0.5mm}
\subsection{Problem Formalization}

Let $\mathcal{D} = \{(x_i, y_i)\}_{i=1}^N$ be a dataset, where $x_i = (t_1, t_2, \dots, t_{T_i})$ represents the $i$-th tweet consisting of $T_i$ tokens, and $y_i \in \{-1, 0, 1\}$ is the associated label. In this scheme, $-1$, $0$, and $1$ correspond to \textquotesingle{}No\textquotesingle{}, \textquotesingle{}Unclear\textquotesingle{}, and \textquotesingle{}Yes\textquotesingle{}, respectively. Defining function $f: \mathcal{X} \to \{-1, 0, 1\}$ that predicts the correct label for each tweet in $\mathcal{X}_{\mathcal{D}} = \{x_i\}_{i=1}^N$, we evaluate $f$ using three metrics: $\alpha$ for accuracy, measured as the fraction of correctly predicted labels relative to a gold standard; $\tau$ for time, representing the combined human and computational time required for complete labeling; and $\kappa$ for cost, reflecting the monetary expense of manual annotation and computational resources.

The labels for this study come from three sources. First, three Amazon Mechanical Turk workers, ${w_1, w_2, w_3}$, each assign a label $y_i^{(w_1)}$, $y_i^{(w_2)}$, or $y_i^{(w_3)}$, with a majority-vote function $g_{\mathrm{AMT}}(\cdot)$ producing the crowd-sourced label $y_i^{(\mathrm{AMT})} = g_{\mathrm{AMT}}(y_i^{(w_1)}, y_i^{(w_2)}, y_i^{(w_3)})$. Second, three domain experts, ${e_1, e_2, e_3}$, independently label each tweet, generating $y_i^{(\mathrm{EXP})} = g_{\mathrm{EXP}}(y_i^{(e_1)}, y_i^{(e_2)}, y_i^{(e_3)})$ through majority voting. We treat $y_i^{(\mathrm{EXP})}$ as the gold standard for measuring $\alpha$, $\tau$, and $\kappa$. Third, an LLM, denoted as $f(\cdot)$, assigns $y_i^{(\mathrm{LLM})}$. We compare $y_i^{(\mathrm{LLM})}$ and $y_i^{(\mathrm{AMT})}$ against $y_i^{(\mathrm{EXP})}$, evaluating accuracy, time, and cost in each case.

\subsection{Data Collection and Preparation}

{\bf Dataset–}We collected tweets and annotations from a public health crowdsourcing repository \cite{shakeri2022physical}, which provides dehydrated tweets for privacy. We rehydrated the tweets using Tweet IDs. From 22,729,110 Canadian tweets, we selected 12,000 across three categories: 5,000 on physical activity, 3,500 on sedentary behavior, and 3,500 on sleep problems.  
Each tweet was labeled by three AMT workers, with the majority vote determining the crowd-sourced label $y_i^{(\mathrm{AMT})}$. Three domain experts provided gold-standard labels $y_i^{(\mathrm{EXP})}$. The labeling scheme (\textquotesingle{}Yes\textquotesingle{}, \textquotesingle{}No\textquotesingle{}, \textquotesingle{}Unclear\textquotesingle{}) indicates whether a tweet represents a recent self-reported instance of the target behavior (Yes = $1$, No = $-1$, Unclear = $0$).

\subsection{Prediction Models Development}
\label{sec:models}

\begin{figure*}
    \centering
    \includegraphics[width=\linewidth]{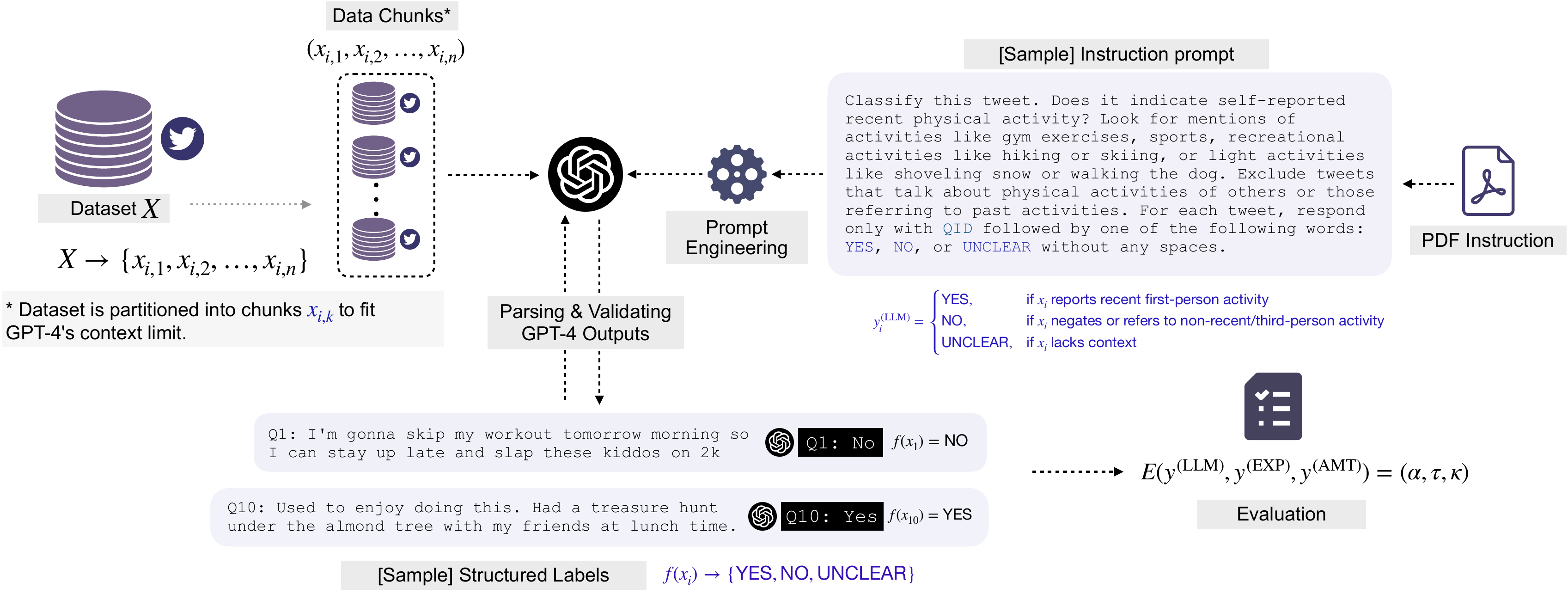}
    \vspace{-5mm}
    \caption{\footnotesize Overview of the LLM-based tweet annotation pipeline implemented in this paper. 
    The dataset \( X \) is partitioned into smaller chunks \( x_{i,k} \) to fit GPT-4\textquotesingle{}s context window. 
    Annotation guidelines are extracted from the instruction PDF and structured into a prompt \( P = g(D) \). 
    GPT-4 generates tweet labels \( y_i^{(\mathrm{LLM})} \), which undergo validation and correction in the output check stage. 
    The final structured labels are evaluated by comparing \( y^{(\mathrm{LLM})} \) with expert (\( y^{(\mathrm{EXP})} \)) and AMT-generated (\( y^{(\mathrm{AMT})} \)) labels using accuracy (\( \alpha \)), time (\( \tau \)), and cost (\( \kappa \)).}
    \label{fig:labeling-example}
     \vspace{-5mm}
\end{figure*}

Figure \ref{fig:labeling-example} presents the implemented pipeline along with an illustrative example. Since each data category (physical activity, sedentary behavior, sleep problems) relies on unique criteria for identifying relevant content, we created separate prompts to guide the labeling function $f(\cdot)$ effectively. We developed four prompt-engineering methods in a sequential manner, where each new prompt method iteratively refined and improved upon the previous one. We assessed the effectiveness of each method by implementing a Python-based pipeline that reads the three category-specific prompts, ingests the tweet data from CSV files, and queries the OpenAI API with the corresponding prompt. After the API returns a raw string containing predicted labels, the pipeline attempts to parse this string into a data frame. If the parsed output fails to match the expected format, or if the set of tweet identifiers in the response does not correspond to those sent to the LLM, the pipeline resends the request until it obtains a consistent output. This approach ensures that $y_i^{(\mathrm{LLM})}$, the label assigned to tweet $x_i$ by the LLM, is captured accurately. The pipeline then concatenates the new labels to form a comprehensive labeled dataset before proceeding to the next batch of tweets. We divide tweets into batches to avoid surpassing the model\textquotesingle{}s context window or output token limit. Finally, the pipeline exports the labeled data frame to a CSV for evaluation.

We began by designing a simple prompt that delineated only the fundamental labeling requirements for each category. However, we observed frequent discrepancies between the prompt\textquotesingle{}s output and the desired format, for instance, invalid tweet identifiers and extraneous explanations. Consequently, we iteratively refined the prompt to address these issues and ensure that the model\textquotesingle{}s output conformed to the specified structure. Subsequently, we employed Prompt Perfect \cite{promptperfect} to optimize our LLM prompts. This tool simulates model outputs and applies internal pattern-matching and validation metrics to compare them against a target structure, automatically refining the prompt by adjusting key parameters and clarifying ambiguous instructions to eliminate discrepancies. Each category\textquotesingle{}s prompt was processed five times, and we selected the variant that achieved the highest labeling accuracy on a sample of 100 tweets.
Since none of the refined prompts yielded satisfactory performance with GPT-4 Turbo, we explored additional strategies to enhance labeling accuracy by reducing the model\textquotesingle{}s inherent randomness. Specifically, we labeled each tweet three times and applied a majority-voting mechanism to the outputs, while also decreasing the temperature parameter of GPT-4 Turbo to limit variability. However, these approaches did not substantially improve accuracy, suggesting that randomness alone was not the primary cause of the model\textquotesingle{}s shortcomings in complex labeling tasks.

We further optimized the pipeline by reintroducing emojis that had been removed during preprocessing, as these symbols sometimes provide additional context required for accurate classification. We also supplied GPT-4 Turbo with the same instructional document used by AMT workers to maintain consistency in labeling guidelines. The model generated refined prompts for each of the three categories, which we applied to batches of 800 or 1000 tweets. Each prompt comprised two components: (1) a header specifying the rules for LLM-based labeling and (2) a list of tweets constrained to remain below the maximum token threshold. A manual review of a subset of responses identified and eliminated problematic outputs, such as verbose justifications or off-topic content.

The final labeled sets produced by our refined pipeline enabled a comparison of $y_i^{(\mathrm{LLM})}$ with $y_i^{(\mathrm{AMT})}$ and $y_i^{(\mathrm{EXP})}$ in terms of accuracy ($\alpha$), time ($\tau$), and cost ($\kappa$). This evaluation established the conditions under which an LLM-based annotation process can effectively augment or replace manual labeling while maintaining high-quality outputs. To ensure replicability and facilitate further research,
the source code of our pipeline
is available on GitHub.\footnote{\href{https://github.com/HIVE-UofT/AMT-vs-LLMs}{GitHub Repository}}

\vspace{-1.5mm}
\section{Results}

Comparisons of labeling accuracy, cost, and time reveal advantages and trade-offs for both GPT-4~Turbo and AMT. Table~\ref{tab:results} lists each approach\textquotesingle{}s accuracy for physical activity, sleep problems, and sedentary behavior, along with overall results. The LLM, denoted by \(y_i^{(\mathrm{LLM})}\), surpasses AMT (denoted by \(y_i^{(\mathrm{AMT})}\)) in two categories, yet underperforms in the sleep problems group. Many sleep-related tweets include medication names or references to non-standard routines that may not explicitly mention sleep, and the model often mislabels these cases. AMT annotators appear to recognize these hidden clues, possibly because human workers can interpret seemingly unrelated words (e.g., brand names of sedatives or abbreviations for prescribed drugs) as indicators of sleep concerns, which improves accuracy in that category.

\begin{table}[h]
    \centering
    \caption{\footnotesize AMT vs. GPT Accuracy in Tweet Labeling}
    \vspace{-2mm}
    \resizebox{\columnwidth}{!}{%
    \begin{tabular}{lccc}
        \toprule
        & \# of Tweets & AMT Accuracy & GPT Accuracy \\
        \midrule
        Physical Activity & 5000 & 79.1\% & 81.0\% \\
        Sleep Problems & 3500 & 69.83\% & 56.14\% \\
        Sedentary Activity & 3500 & 64.69\% & 73.31\% \\
        \midrule
        Total & 12000 & 70.21\% & 71.51\% \\
        \bottomrule
    \end{tabular}
}
    \label{tab:results}
     \vspace{-3mm}
\end{table}

Table~\ref{tab:consolidated_confusion} consolidates the confusion matrices for \(y_i^{(\mathrm{LLM})}\) and \(y_i^{(\mathrm{AMT})}\), using the expert-derived \(y_i^{(\mathrm{EXP})}\) as the actual label. The diagonal cells, highlighted in light purple, show correct classifications, while the off-diagonal cells represent misclassifications. The LLM frequently commits errors when tweets contain overlapping references, for example, a single post mentioning both a short workout and prolonged sedentary periods. AMT workers also show confusion in certain borderline cases, but crowd workers seem more likely to interpret lesser-known medication names or partial references that hint at sleep disturbances.

\begin{table*}[ht]
\centering
\caption{\footnotesize Confusion Matrices for \(y_i^{(\mathrm{LLM})}\) and \(y_i^{(\mathrm{AMT})}\), 
with \(y_i^{(\mathrm{EXP})}\) used as the actual label. 
Diagonal cells indicate correct predictions.}
\label{tab:consolidated_confusion}
\renewcommand{\arraystretch}{1.2}
\resizebox{.7\textwidth}{!}{%
\begin{tabular}{l|ccc|ccc}
\toprule
\multirow{2}{*}{\textbf{Actual} \(y_i^{(\mathrm{EXP})}\)} 
& \multicolumn{3}{c|}{\textbf{GPT: }\(y_i^{(\mathrm{LLM})}\)} 
& \multicolumn{3}{c}{\textbf{AMT: }\(y_i^{(\mathrm{AMT})}\)} \\
& \textbf{Yes} & \textbf{No} & \textbf{Unclear} & \textbf{Yes} & \textbf{No} & \textbf{Unclear} \\
\midrule
\textbf{Yes} 
& \cellcolor{hivePurple!20}356 (2.97\%) & 1826 (15.21\%) & 5 (0.04\%) & \cellcolor{hivePurple!20}1471 (12.26\%) & 696 (5.8\%) & 20 (0.17\%) \\
\textbf{No} 
& 415 (3.46\%) & \cellcolor{hivePurple!20}8222 (68.52\%) & 5 (0.04\%) & 1688 (14.07\%) & \cellcolor{hivePurple!20}6869 (57.24\%) & 85 (0.71\%) \\
\textbf{Unclear} 
& 115 (0.96\%) & 1053 (8.78\%) & \cellcolor{hivePurple!20}3 (0.03\%) & 452 (3.77\%) & 689 (5.74\%) & \cellcolor{hivePurple!20}30 (0.25\%) \\
\bottomrule
\end{tabular}
}
 \vspace{-5mm}
\end{table*}

Adjusting the temperature parameter in GPT-4~Turbo to reduce randomness did not enhance the overall accuracy and, in some cases, lowered it. This finding implies that more deterministic sampling often leads to overconfidence in the model\textquotesingle{}s most probable choice, causing oversights in messages that require subtle interpretation. AMT workers can draw on real-world knowledge to interpret ambiguous text, which likely improves their performance when tweets contain brand names or abbreviations for sleep-related medications. Meanwhile, the LLM\textquotesingle{}s misclassification errors tend to occur when a tweet contains multiple overlapping signals or partial clues that do not strongly match any single label category.

Cost and time point to clear advantages for GPT-4~Turbo on larger datasets. GPT-4~Turbo charges \$0.01 for each 1,000 input tokens and \$0.03 for each 1,000 output tokens, leading to a total labeling cost of approximately \textbf{\$6} for this dataset, even when factoring in multiple requests to address format mismatches. AMT labeling operated at a rate of \$0.03–\$0.05 per page, where each page contained three labeling questions and one quality-check example.  Labeling 12,000 tweets used in this study required 4,000 pages, which cost between \textbf{\$120} and \textbf{\$200}, depending on the pay rate. AMT costs rise further with incentives to enhance annotation quality and reduce random responses.

Time measurements further support the economic advantages of an LLM-based workflow. GPT-4~Turbo labeled 800 tweets in about four minutes using Python~3.11 on Google~Colab, resulting in a total labeling time of around one hour for the 12,000 tweets. AMT relied on crowd worker availability and task acceptance, as well as qualification steps and quality checks to filter out low-effort labeling. This process for the dataset used in this study took approximately one week to 10~days to finalize and clean the labeled dataset \cite{shakeri2022physical, shakeri2022crowdsourcing}. The observed speed and cost efficiency of LLM-based pipelines suggest their potential for large-scale text annotation; however, additional methodological refinements are necessary to address ambiguities and ensure robust performance.

\vspace{-2mm}
%%%%%%%%%%%%%%%%%%%%%%%%%%%%%%%%%%%%%%%%%%%%%%%%%%%%%%%%%%%%%%%%%%%%%%%%%%%%%%%%
\section{Discussion and Conclusion}

LLMs such as GPT-4 handle zero-shot labeling with notable efficiency \cite{brown2020language,radford2019language}, making them valuable for large-scale public health surveillance. When assigning \(y_i^{(\mathrm{LLM})}\) to tweets spanning topics like physical activity, sedentary behavior, and sleep problems, the model rapidly processes data at minimal cost. In scenarios where tweets contain clear references, GPT-4 reaches accuracy levels close to AMT (\(y_i^{(\mathrm{AMT})}\)), indicating that a hybrid workflow using LLMs for common cases and human review for ambiguities may improve speed while preserving accuracy. 

The sleep problems category illustrates the difficulty of capturing indirect cues, such as cryptic medication names or implied symptoms. GPT-4 often treats such posts as definitively positive or negative, overlooking \textquotesingle{}Unclear\textquotesingle{} labels. This gap arises from generalized language patterns that fail to recognize complex health contexts, reflecting observations in prior work. Human annotators (\(y_i^{(\mathrm{AMT})}\)) interpret these subtle markers more accurately, aligning more closely with expert assessments (\(y_i^{(\mathrm{EXP})}\)). These findings support calls for targeted fine-tuning or domain adaptation, particularly in areas where language norms differ substantially from mainstream usage.

{\bf Challenges in LLM-Based Labeling.} 
The low usage of the \textquotesingle{}Unclear\textquotesingle{} label by GPT-4 Turbo aligns with previous studies indicating that models prefer more definitive tokens once a probable decision boundary is identified \cite{zhong2024evaluation}. Attempts to lower the temperature parameter did not correct the underuse of \textquotesingle{}Unclear\textquotesingle{} and, in some instances, reduced accuracy. This tendency suggests that more deterministic outputs often cause overconfidence in the most likely label, skipping fine-grained signals that occur in tweets mentioning prescriptions or activities with multiple interpretations. Crowd workers, on the other hand, showed more variation in labeling, reflecting human ability to handle multi-faceted clues. Public health data often contain irregular personal statements, requiring interpretation of incomplete language for accurate classification.

{\bf Limitations.} 
Several limitations should be noted. First, our reliance on domain experts \(\{e_1, e_2, e_3\}\) to derive gold labels \(y_i^{(\mathrm{EXP})}\) introduces subjectivity, especially given the ambiguity in some tweets. We mitigated this by using three independent experts and majority voting to reduce individual bias, though consensus remains only a proxy for absolute truth. Second, our exclusive use of GPT-4, selected for its robust zero-shot performance, limits the evaluation scope, as other LLMs might offer different advantages. Future work should consider a broader range of models. Third, potential biases in our dataset may affect generalizability, as the tweets might not represent the full spectrum of social media content. We addressed this by selecting categories with varying contextual complexity (physical activity, sedentary behavior, and sleep problems) to enhance robustness.

{\bf Potential Directions.} Future work could explore parameter-efficient training methods, such as Low-Rank Adaptation (LoRA) \cite{hu2021loralowrankadaptationlarge}, to enhance LLMs for specialized medical topics. Fine-tuning with a small set of domain-specific examples may help the model to detect references to medications or private routines. Further, exploring models that emphasize reasoning tokens \cite{zhong2024evaluation} could improve performance in categories where language cues are less pronounced. These approaches offer a promising route toward an integrated pipeline that combines speed and reliability for large-scale public health surveillance at sustainable costs.

\vspace{-3mm}
%%%%%%%%%%%%%%%%%%%%%%%%%%%%%%%%%%%%%%%%%%%%%%%%%%%%%%%%%%%%%%%%%%%%%%%%%%%%%%%%
\section*{ACKNOWLEDGMENT}
This research is supported by the Vector Scholarship in Artificial Intelligence from the Vector Institute, the Artificial Intelligence for Public Health (AI4PH) Trainee Award funded by the Canadian Institutes of Health Research (CIHR), and the Natural Sciences and Engineering Research Council of Canada through the Canada Research Chairs  program.

%%%%%%%%%%%%%%%%%%%%%%%%%%%%%%%%%%%%%%%%%%%%%%%%%%%%%%%%%%%%%%%%%%%%%%%%%%%%%%%%

\vspace{-1mm}

\printbibliography

\end{document}